%% file: fquad.tex
\documentclass{article}

\usepackage{two_columns}

\usepackage[utf8]{inputenc} 
\usepackage[T1]{fontenc}    
\usepackage{hyperref}   
\usepackage{url}       
\usepackage{booktabs}       
\usepackage{amsfonts}      
\usepackage{nicefrac}   
\usepackage{microtype}      
\usepackage{lipsum}
\usepackage{amsmath}
\usepackage{float}
\usepackage{xcolor}
\definecolor{curveblue}{rgb}{0.1725490196,0.74901960784,0.96862745098}
\definecolor{curvered}{rgb}{0.98823529411, 0.34117647058, 0.05882352941}
\usepackage{natbib}
\usepackage{makecell}
\newcommand{\specialcell}[2][l]{%
  \begin{tabular}[#1]{@{}l@{}}#2\end{tabular}}
\usepackage{arydshln}

\usepackage{graphics}
\usepackage{graphicx}
\usepackage{tikz}
\usepackage{todonotes}

\usepackage{multirow}
\usepackage{pgfplots}
\usepackage{appendix}
\usepackage{textcomp}
\usepackage{gensymb}

\definecolor{internationalkleinblue}{rgb}{0.0, 0.18, 0.65}
\hypersetup{
    colorlinks=true,
    linkcolor=internationalkleinblue,
    filecolor=magenta,      
    urlcolor=internationalkleinblue,
}

\title{FQuAD: French Question Answering Dataset}

\author{
    \normalsize Martin d'Hoffschmidt \hspace{0.3cm} \normalsize Wacim Belblidia \hspace{0.3cm} \normalsize Tom Brendlé \hspace{0.3cm} \normalsize Quentin Heinrich \\
    \normalsize \textsc{Illuin Technology} \\
    \normalsize Paris, France\\
    \normalsize \texttt{\{martin, wacim, tom, quentin\}@illuin.tech} \\
    \AND
    \normalsize {\bf Maxime Vidal} \hspace{0.3cm} \\
    \normalsize \textsc{ETH Zurich}\\
    \normalsize \texttt{mvidal@student.ethz.ch} \\
}

\pgfplotsset{compat=1.14}
\begin{document}

\maketitle

\begin{abstract}
\input{content/abstract.tex}
\end{abstract}

\section{Introduction}
\label{section:introduction}
\input{content/introduction.tex}

\section{Related Work}
\label{section:related_work}
\input{content/related_work.tex}

\section{Dataset Collection}
\label{section:dataset_collection}
\input{content/dataset_collection.tex}

\section{Dataset Analysis}
\label{section:dataset_analysis}
\input{content/dataset_analysis.tex}

\section{Dataset Evaluation}
\label{section:dataset_evaluation}
\input{content/dataset_evaluation.tex}

\section{Experiments}
\label{section:experiments}
\input{content/experiments.tex}

\section{Results}
\label{section:results}
\input{content/results.tex}

\section{Discussion}
\label{section:discussion}
\input{content/discussion.tex}

\section{Conclusion}
\label{section:conclusion}
\input{content/conclusion.tex}

\section{Acknowledgments}
\input{content/acknowledgement.tex}

\bibliography{references}

\clearpage
\appendix

\input{output_examples.tex}

\end{document}

%% file: content/abstract.tex
Recent advances in the field of language modeling have improved state-of-the-art results on many Natural Language Processing tasks.
Among them, Reading Comprehension has made significant progress over the past few years.
However, most results are reported in English since labeled resources available in other languages, such as French, remain scarce.
In the present work, we introduce the \textbf{F}rench \textbf{Qu}estion \textbf{A}nswering \textbf{D}ataset (FQuAD).
FQuAD is a French Native Reading Comprehension dataset of questions and answers on a set of Wikipedia articles that consists of 25,000+ samples for the 1.0 version and 60,000+ samples for the 1.1 version.
We train a baseline model which achieves an F1 score of 92.2 and an exact match ratio of 82.1 on the test set. 
In order to track the progress of French Question Answering models we propose a leader-board and we have made the 1.0 version of our dataset freely available at \url{https://illuin-tech.github.io/FQuAD-explorer/}.

%% file: content/introduction.tex
Current progress in language modeling has led to increasingly successful results on various Natural Language Processing (NLP) tasks such as Part of Speech Tagging (PoS), Named Entity Recognition (NER) and Natural Language Inference (NLI).
Large amounts of unstructured text data available for most languages have facilitated the development of language models.
Therefore, the releases of language specific models in Japanese, Chinese, German and Dutch \citep{Vries2019BERTjeAD}, amongst other languages, are now thriving as well as multilingual models \citep{multilingual-bert} and \citep{xlmr}.
Recently, two French language models, CamemBERT \citep{camembert} and FlauBERT \citep{flaubert}, were released.

However, language specific datasets are costly and difficult to collect. 
This is especially the case with the Reading Comprehension task \citep{richardson-etal-2013-mctest}.
On one hand, numerous English datasets have been released such as SQuAD1.1 \citep{rajpurkar-etal-2016-squad}, SQuAD2.0 \citep{rajpurkar-squad-v2} or CoQA \citep{CoQA} that fostered important and impressive progress for English Question Answering models over the past few years.
On the other hand, the lack of native language annotated datasets apart from English is one of the main reasons why the development of language specific Question Answering models is slower.
This is namely the case for French.

To tackle this problem, substantial efforts have been carried out recently to come up with native Reading Comprehension datasets in for instance Korean \citep{lim2019korquad10}, Russian \citep{efimov2019sberquad} and Chinese \citep{cui-etal-2019-span}. 
A more appealing solution in terms of cost and time efficiency relies on leveraging the advances in Neural Machine Translation (NMT) to translate the English datasets in target languages to fine-tune the language model on the translated dataset.
This is for instance the case of \citeauthor{spanishsquad} where SQuAD1.1 is translated in Spanish in order to train a multilingual model to answer Spanish questions.
An alternative is proposed by \citep{xquad} and \citep{mlqa} where the authors provide a cross-lingual evaluation benchmark to enhance the development of cross-lingual Question Answering models that can transfer to a target language without requiring training data in that language.
However, in both cases, the reported performances fail to reach English comparable results on other languages.

In order to fill the gap for the French language, we release a French Reading Comprehension dataset similar to SQuAD1.1.
The dataset consists of French native questions and answers samples annotated by a team of university students. 
The dataset comes in two versions. 
First FQuAD1.0, containing over 25,000+ samples. 
Second, FQuAD1.1 containing over 60,000+ samples. 
The 35k+ additional samples have been annotated with more demanding guidelines to strengthen complexity of the data and model to make the task harder.
More specifically, the training, development and test sets of FQuAD1.0 contain respectively 20703, 3188 and 2189 samples.
And the training, development and test sets of FQuAD1.1 contain respectively 50741, 5668 and 5594 samples.

In order to evaluate the FQuAD dataset, we perform various experiments by fine-tuning BERT based Question Answering models on both versions of the FQuAD dataset.
Our experiments cover not only the recently released French pre-trained language models CamemBERT \citep{camembert} and FlauBERT \citep{flaubert} but also multilingual models such as mBERT \citep{multilingual-bert}, XLM-RoBERTa \citep{xlmr} in order to better understand how multilingual models perform on native datasets other than English.

Finally, we perform two types of cross-lingual Reading Comprehension experiences.
First, we evaluate the performance of the zero-shot cross-lingual transfer learning approach as stated in \citep{xquad} and \citep{mlqa} on our newly obtained native French dataset. 
Second, we evaluate the performance of the translation approach by fine-tuning models on the French translated version of SQuAD1.1.
The results of these two experiments help to better understand how the two cross-lingual approaches actually perform on a native dataset.

%% file: content/related_work.tex
The Reading Comprehension task (RC) \citep{richardson-etal-2013-mctest}, \citep{rajpurkar-etal-2016-squad} attempts to solve the Question Answering (QA) problem by finding the text span in one or several documents or paragraphs that answers a given question \citep{nlpprogress}.

\subsection{Reading Comprehension in English}

Many Reading Comprehension datasets have been built in English.
Among them SQuAD1.1 \citep{rajpurkar-etal-2016-squad}, then later SQuAD2.0 \citep{rajpurkar-squad-v2} has become one of the major reference dataset for training question answering models.
Later, similar initiatives such as NewsQA \citep{NewsQA}, CoQA \citep{CoQA}, QuAC \citep{quac}, HotpotQA \citep{yang2018hotpotqa} have broadened the research area for English Question Answering. 

These datasets are similar but each of them introduces its own subtleties.
For instance, SQuAD2.0 \citep{rajpurkar-squad-v2} develops unanswerable adversarial questions.
CoQA \citep{CoQA} focuses on Conversation Question Answering in order to measure the ability of algorithms to understand a document and answer series of interconnected questions that appear in a conversation.
QuAC \citep{quac} focuses on Question Answering in Context developed for Information Seeking Dialog (ISD).
The benchmark established by \citep{qa-benchmark} offers a qualitative comparison of these datasets.
Finally, HotPotQA\citep{yang2018hotpotqa} attempts to extend the Reading Comprehension task to more complex reasoning by introducing Multi Hop Questions (MHQ) where the answer must be found among multiple documents.

\subsection{Reading Comprehension in other languages}

Native Reading Comprehension datasets other than English remain rare.
Among them, some initiatives have been carried out in Chinese, Korean and Russian and all of them have been built in a similar way to SQuAD1.1.
The SberQuAD dataset \citep{efimov2019sberquad} is a Russian native Reading Comprehension dataset and is made up of 50,000+ samples.
The CMRC 2018 \citep{cui-etal-2019-span} dataset is a Chinese native Reading Comprehension dataset that gathers 20,000+ question and answer pairs.
The KorQuAD dataset \citep{lim2019korquad10} is a Korean native Reading Comprehension dataset that is made up of 70,000+ samples.
Note that following our work, the PIAF project \citep{piaf} has released a native French Dataset of 3835 question and answer pairs.

As language specific datasets are costly and challenging to obtain, an alternative consists in developing cross-lingual models that can transfer to a target language without requiring training data in that language \citep{mlqa}.
It has indeed been shown that these unsupervised multilingual models generalize well in a zero-shot cross-lingual setting \citep{xquad}.
For this reason, cross-lingual Question Answering has recently gained traction and two cross-lingual benchmarks have been released, i.e XQuAD \citep{xquad} and MLQA \citep{mlqa}.
The XQuAD dataset \citep{xquad} is obtained by translating 1190 question and answer pairs from the SQuAD1.1 development set by professionals translators in 10 foreign languages. 
The MLQA dataset \citep{mlqa} consists of over 12000 question and answer samples in English and 5000 samples in 6 other languages such as Arabic, German and Spanish. 
Note that the two aforementioned datasets do not cover French.

Another alternative consists in translating the training dataset into the target language and fine-tuning a language model on the translated dataset.
This is namely the case of \citep{spanishsquad} where the authors develop a specific translation method called Translate-Align-Retrieve (TAR) to translate the English SQuAD1.1 dataset into Spanish.
The resulting Spanish SQuAD1.1 dataset is used to fine-tune a multilingual model that reaches a performance of respectively 68.1/48.3\% F1/EM and 77.6/61.8\% F1/EM on MLQA cross-lingual benchmark \citep{mlqa} and XQuAD\citep{xquad}.
Note that a similar approach has been adopted for French and Japanese in \citep{french-translated-squad} and \citep{mBERT-qa-french}.
In \citep{mBERT-qa-french} a multilingual BERT is trained on English texts of SQuAD1.1, and evaluated on the small translated \citeauthor{french-translated-squad} French corpus. 
This set-up reaches a promising score of 76.7/61.8 \% F1/EM.

\subsection{Language modeling for Reading Comprehension}

Increasingly efficient language models have been released recently such as GPT-2 \citep{gpt2}, BERT \citep{bert}, XLNet \citep{xlnet} and RoBERTa \citep{roberta}.
They have indeed disrupted the Reading Comprehension task and most of NLP fields: pre-training a language model on a generic corpus, eventually fine-tuning it on a domain specific corpus and then training it on a downstream task is the de facto state-of-the-art approach for optimizing both performances and annotated data volumes \citep{bert}, \citep{roberta}.
For instance, the top performing models on the SQuAD1.1 and SQuAD2.0 leader-boards\footnote{\url{rajpurkar.github.io/SQuAD-explorer}} are essentially transformer based models. 
Unfortunately, the aforementioned models are pretrained on English corpora and their use for French is therefore limited.

Multilingual models pre-trained on large multi-lingual datasets attempt to alleviate the language specific shortcoming characteristic of the former models such as \citep{xlm}, \citep{multilingual-bert} and more recently XLM-R \citep{xlmr}.
It has been shown in \citep{xlmr}, \citep{xquad} and \citep{mlqa} that multilingual models are flexible and perform reasonably well on other languages than English.
However, they do not appear to perform better than specific language models \citep{mlqa}.

Regarding French, few resources were available until recently. 
First, the CamemBERT models \citep{camembert} were trained on 138 GB of French text from the Oscar dataset \citep{oscar_dataset}.
Second, the FlauBERT models \citep{flaubert} were trained on 71 GB of text.
Note that both models were pre-trained with the Masked Language Modeling task only \citep{camembert}, \citep{flaubert}.
Both models reach similar performances on French NLP tasks such as PoS, NER and NLI. 
However, their performance has not yet been evaluated on the Reading Comprehension task as no French dataset is available.

\begin{table}[ht]
    \centering
    \begin{tabular}{l c l}
        Dataset & Language & Size \\
        \hline
        SQuAD1.1 & English & 100k+ \\
        SQuAD2.0 & English & 150k \\
        NewsQA   & English & 100k+ \\
        CoQA     & English & 127k+ \\
        QuAC     & English & 98k+ \\
        HotpotQA & English & 113k+ \\
        \hline
        KorQuAD & Korean & 70k+ \\
        \hline
        SberQuAD & Russian & 50k+ \\
        \hline
        CMR-2018 & Chinese & 20k+ \\
        \hline
        FQuAD1.0 & French & \textbf{25k+} \\
        FQuAD1.1 & French & \textbf{60k+} \\
        PIAF & French & 3835 \\
        \hline
    \end{tabular}
    \caption{Benchmark of existing Reading Comprehension datasets, including FQuAD.}
    \label{tab:dataset-benchmark}
\end{table}

Finally, Table \ref{tab:dataset-benchmark} lists some of the available datasets along with the number of samples they contain\footnote{\url{https://nlpprogress.com/english/question_answering.html}}. 
By means of comparison, Table \ref{tab:dataset-benchmark} also includes FQuAD, whose collection is presented in the upcoming sections.

%% file: content/dataset_collection.tex
The collection procedure for our dataset follows the same standards and guidelines as SQuAD1.1 \citep{rajpurkar-etal-2016-squad}.
First, paragraphs among diverse articles are collected.
Second, question and answer pairs are crowd-sourced on the collected paragraphs.
Third, additional answers are collected for the development and test sets.
The Dataset Collection was conducted in two distinct steps: the first one resulted in FQuAD1.0 with 25k+ question and answer pairs, and the second one resulted in FQuAD1.1 with 60k+ question and answer pairs.

\subsection{Paragraphs collection}

A set of 1,769 articles are collected from the French Wikipedia page referencing quality articles \footnote{\url{https://fr.wikipedia.org/wiki/Catégorie:Article_de_qualité}}.
From this set, a total of 145 articles are randomly sampled to build the FQuAD1.0 dataset.
Also, 181 additional articles are randomly sampled to extend the dataset to FQuAD1.1. resulting in a total of 326 articles.
Among them, articles are randomly assigned to the training, development and test sets.
The training, development and test sets for FQuAD1.0 are respectively made up of 117, 18 and 10 articles.
For the FQuAD1.1 dataset, they are respectively made up of 271, 30 and 25 articles.
Note that train, development, test split is performed at the article level in order to avoid any possible biases.

The paragraphs that are at least 500 characters long are kept for each article, similarly to \citep{rajpurkar-etal-2016-squad}.
This technique results in 4951, 768 and 523 paragraphs for respectively the training, development and test sets of FQuAD1.0.
For FQuAD1.1, the number of collected paragraphs for the same sets are respectively 12123, 1387 and 1398.

\subsection{Question and answer pairs collection}

A specific annotation platform was developed to collect the question and answer pairs.
The workers were hired in collaboration with the Junior Enterprise of CentraleSupélec \footnote{\url{https://juniorcs.fr/en/}}.

\begin{figure}[ht]
    \centering
    \includegraphics[width=0.47\textwidth]{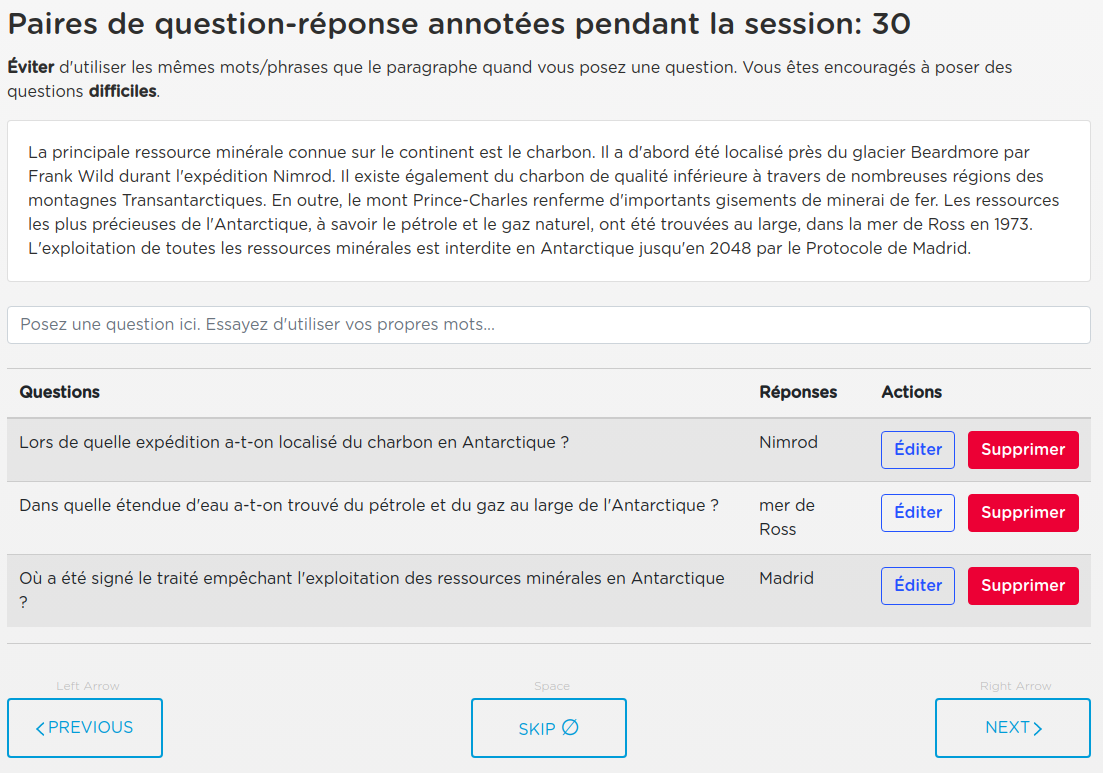}
    \caption{The interface used to collect the question/answers encourages workers to write difficult questions.}
\end{figure}

The guidelines for writing question and answer pairs for each paragraph are the same as for SQuAD1.1 \citep{rajpurkar-etal-2016-squad}.
First, the paragraph is presented to the student on the platform and the student reads it.
Second, the student thinks of a question whose answer is a span of text within the context. 
Third, the student selects the smallest span in the paragraph which contains the answer. 
The process is then repeated until 3 to 5 questions are generated and correctly answered. 
The students were asked to spend on average 1 minute on each question and answer pair.
This amounts to an average of 3-5 minutes per annotated paragraph.
Final dataset metrics are shared in table \ref{tab:fquad1.1}.

\subsection{Additional answers collection}

Additional answers are collected to decrease the annotation bias similarly to \citep{rajpurkar-etal-2016-squad}.
For each question in the development and test sets, two additional answers are collected, resulting in three answers per question for these sets.
The crowd-workers were asked to spend on average 30 seconds to answer each question.

For the same question, several answers may be correct: for instance the question \textit{Quand fut couronné Napoléon ?} would have several possible answers such as \textit{mai 1804}, \textit{en mai 1804} or \textit{1804}.
As all those answers are admissible, enriching the test set with several annotations for the same question, with different annotators, is a way to decrease annotation bias. 
The additional answers are useful to get an indication of the human performance on FQuAD. 

\subsection{FQuAD1.0}

The results for the first annotation process resulting in the FQuAD1.0 dataset are reported in table \ref{tab:fquad1.0}. 
The number of collected question and answer pairs amounts to 26108.
Diverse analysis to measure the difficulty of the resulting dataset are performed as described in the next section. 
A complete annotated paragraph is displayed in figure \ref{fig:annotation-sample}.
\input{examples/fquad_example.tex}

\begin{table}[ht]
    \centering
    \resizebox{0.47\textwidth}{!}{
    \begin{tabular}{l r r r}
        Dataset         & Articles  & Paragraphs    & Questions \\
        \hline
        Train           & 117       & 4921          & 20731 \\
        Development      & 18        & 768           & 3188 \\
        Test            & 10        & 532           & 2189 \\
        \hline
    \end{tabular}}
    \caption{The number of articles, paragraphs and questions for FQuAD1.0}
    \label{tab:fquad1.0}
\end{table}

\subsection{FQuAD1.1}

The first dataset is extended with additional annotation samples to build the FQuAD1.1 dataset reported in table \ref{tab:fquad1.1}. 
The total number of questions amounts to 62003.
The FQuAD1.1 training, development and test sets are then respectively composed of 271 articles (83\%), 30 (9\%) and 25 (8\%).
The difference with the first annotation process is that the workers were specifically asked to come up with complex questions by varying style and question types in order to increase difficulty.
The additional answer collection process remains the same.

\begin{table}[ht]
    \centering
    \resizebox{0.47\textwidth}{!}{
    \begin{tabular}{l r r r}
        Dataset         & Articles  & Paragraphs    & Questions \\
        \hline
        Train           & 271       & 12123         & 50741 \\
        Development      & 30        & 1387          & 5668  \\
        Test            & 25        & 1398          & 5594  \\
        \hline
    \end{tabular}}
    \caption{The number of articles, paragraphs and questions for FQuAD1.1}
    \label{tab:fquad1.1}
\end{table}

\subsection{Adversarial samples}

The present dataset does not contain adversarial samples as in SQuAD2.0 by \citep{rajpurkar-squad-v2}.
However, this will hopefully be released in a future version of the dataset.


%% file: examples/fquad_example.tex
\begin{figure}[ht]
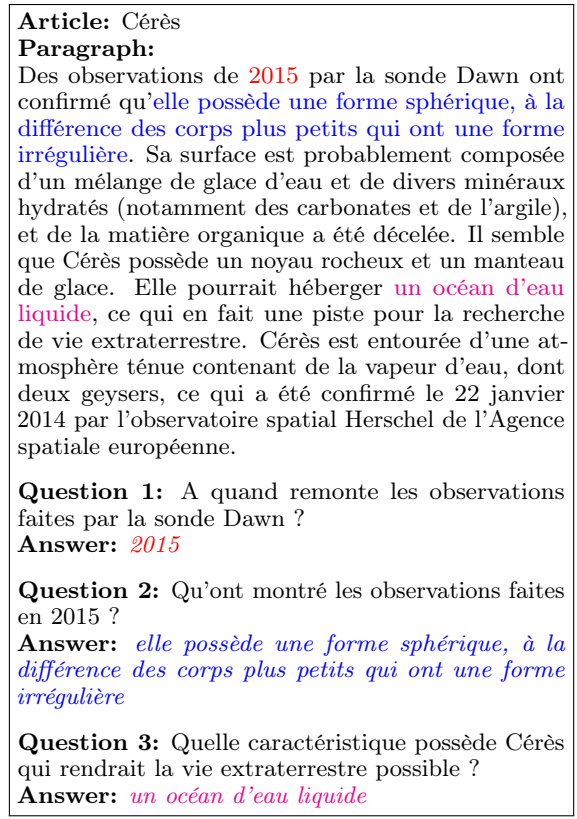

  \noindent\fbox{%
    \parbox{0.45\textwidth}{
  \footnotesize
    \textbf{Article:} Cérès

    \textbf{Paragraph:}

    Des observations de \textcolor{red}{2015} par la sonde Dawn ont confirmé qu'\textcolor{blue}{elle possède une forme sphérique, à la différence des corps plus petits qui ont une forme irrégulière}. Sa surface est probablement composée d'un mélange de glace d'eau et de divers minéraux hydratés (notamment des carbonates et de l'argile), et de la matière organique a été décelée. Il semble que Cérès possède un noyau rocheux et un manteau de glace. Elle pourrait héberger \textcolor{magenta}{un océan d'eau liquide}, ce qui en fait une piste pour la recherche de vie extraterrestre. Cérès est entourée d'une atmosphère ténue contenant de la vapeur d'eau, dont deux geysers, ce qui a été confirmé le 22 janvier 2014 par l'observatoire spatial Herschel de l'Agence spatiale européenne.

    \vspace{0.1in}

    \textbf{Question 1:} 
    A quand remonte les observations faites par la sonde Dawn ?

    \textbf{Answer:} \textit{\textcolor{red}{2015}}
    
    \vspace{0.1in}

    \textbf{Question 2:} 
    Qu'ont montré les observations faites en 2015 ?

    \textbf{Answer:} \textit{\textcolor{blue}{elle possède une forme sphérique, à la différence des corps plus petits qui ont une forme irrégulière}}
    
    \vspace{0.1in}

    \textbf{Question 3:} 
    Quelle caractéristique possède Cérès qui rendrait la vie extraterrestre possible ?

    \textbf{Answer:} \textit{\textcolor{magenta}{un océan d'eau liquide}}

  }}
  \caption{Question answer pairs for a sample passage in FQuAD}
  \label{fig:annotation-sample}
\end{figure}

%% file: content/dataset_analysis.tex
In order to understand the diversity of the dataset, we perform various analysis.
First, a mix of PoS-tagging and patterns is used to analyse the frequency of different kinds of answers (see table \ref{tab:dataset_analysis_answer}).
Second, a keyword based approach is used to analyse the frequency of the corresponding questions (see table \ref{tab:dataset_analysis_question}).
Finally, we present the result of our analysis on the question-answer differences.

\subsection{Answer analysis}

To analyse the collected answers, a combination of rule-based regular expressions and entity extraction using spaCy \citep{spacy2} are used.
First, a set of regular expression rules are applied to isolate \texttt{dates} and other \texttt{numerical} answers.
Second, \texttt{person} and \texttt{location} entities are extracted using Named Entity Recognition. 
Third, a rule based approach is adopted to extract the remaining \texttt{proper nouns}. 
Finally, the remaining answers are labeled into \texttt{common noun, verb} and \texttt{adjective} phrases, or \texttt{other} if no labels were found. 
Answer type distribution is shown in table \ref{tab:dataset_analysis_answer}.

\begin{table}[ht]
    \centering
    \resizebox{0.47\textwidth}{!}{%
    \begin{tabular}{l c l}
        Answer type             & Freq [\%] & Example \\
        \hline
            Common noun         & 26.6      & rencontres \\
            Person              & 14.6        & John More \\
            Other proper nouns  & 13.8      & Grand Prix d'Italie\\
            Other numeric       & 13.6      & 1,65 m \\
            Location            & 14.1      & Normandie \\
            Date                & 7.3       & 1815 \\
            Verb                & 6.6       & être dépoussiéré \\
            Adjective           & 2.6       & méprisant, distant et sec \\
            Other               & 0.9       & gimmick \\
        \hline
    \end{tabular}}
    \caption{Answer type by frequency for the development set of FQuAD1.1}
    \label{tab:dataset_analysis_answer}
\end{table}

\subsection{Question analysis}

The second analysis aims at understanding the question types of the dataset.
The present analysis is performed rule-based only.
Table \ref{tab:dataset_analysis_question} first demonstrates that the annotation process issued a wide range of question types, underlining the fact that \textit{What (que)} represents almost half (47.8\%) of the corpus. 
This important proportion may be explained by this formulation encompassing both the English \textit{What} and \textit{Which}, as well as a possible natural bias in the annotators way of asking questions. 
Our intuition is that this bias is the same during inference, as it originates from native French structure.

\begin{table}[ht]
    \centering
    \resizebox{0.47\textwidth}{!}{%
    \begin{tabular}{l c l}
        Question            & Freq [\%] & Example \\
        \hline
            What (que)      & 47.8      & Quel pays parvient à ... \\
            Who             & 12.2      & Qui va se marier bientôt ?\\
            Where           & 9.6       & Où est l'échantillon ...\\
            When            & 7.6       & Quand a eu lieu la ...\\
            Why             & 5.3       & Pourquoi l'assimile ...\\
            How             & 6.8       & Comment est le prix ...\\
            How many        & 5.6       & Combien d'albums ...\\
            What (quoi)     & 4.1       & De quoi est faite la ...\\
            Other           & 1       & Donner un avantage de ...\\
        \hline
    \end{tabular}}
    \caption{Question type by frequency for the development set of FQuAD1.1}
    \label{tab:dataset_analysis_question}
\end{table}

\subsection{Question-answer differences}

The difficulty in finding the answer given a particular question lies in the linguistic variation between the two. 
This can come in different ways, which are listed in table \ref{tab:dataset_analysis_difficulty}
The categories are taken from \citep{rajpurkar-etal-2016-squad}: \textit{Synonymy} implies key question words are changed to a synonym in the context; \textit{World knowledge} implies key question words require world knowledge to find the correspondence in the context; \textit{Syntactic variation} implies a difference in the structure between the question and the answer; \textit{Multiple sentence reasoning} implies knowledge requirement from multiple sentences in order to answer the question.
We randomly sampled 6 questions from each article in the development set and manually labeled them.
Note that samples can belong to multiple categories.

\begin{table*}[ht]
    \small
    \centering
   \resizebox{\textwidth}{!}{%
    \begin{tabular}{l l c}
        Reasoning & Example & Frequency \\
        \hline
        \rule{0pt}{0.35 in} Synonymy &  \specialcell{Question: Quel est le sujet \textbf{principal} du film ? \vspace{0.1in}\\ Context: Le sujet \textbf{majeur} du film est le \textit{conflit de Rick Blaine entre l'amour et} \\ \textit{la vertu} : il doit choisir entre... } \vspace{0.05in}& 35.2 \% \\
        \hline
         \rule{0pt}{0.43in}World knowledge & \specialcell{ Question: Quand John Gould a-t-il décrit la nouvelle \textbf{espèce d'oiseau} ? \vspace{0.1in}\\  Context: \textbf{E. c. albipennis} décrite par John Gould en \textit{1841}, se rencontre dans \\le nord du Queensland, l'ouest du golfe de Carpentarie dans le Territoire du No- \\ rd  et dans le nord de l'Australie-Occidentale.} \vspace{0.05in}& 11.1 \% \\
        \hline
        \rule{0pt}{0.35in} Syntactic variation & \specialcell{Question: Combien d'\textbf{auteurs ont parlé de la merveille du monde de Babylone} ? \vspace{0.1in}\\ Context: Dès les premières campagnes de fouilles, on chercha \textbf{la « merveille} \\ \textbf{du monde » de Babylone : les Jardins suspendus décrits par }\textit{cinq} auteurs...} \vspace{0.05in}& 57.4 \% \\
        \hline
        \rule{0pt}{0.44in} Multiple sentence reasoning & \specialcell{Question: Qu'est ce qui rend la situation de menace des cobs précaire ? \vspace{0.1in}\\ Context:  En 1982, les chercheurs en concluent que \textbf{le cob normand} est victime \\ de consanguinité, de dérive génétique et de la disparition de ses structures de \\ coordination. \textit{L'âge avancé de ses éleveurs} rend \textbf{sa} situation précaire.} & 17.6 \vspace{0.05in} \% \\
        \hline
    \end{tabular}
    }
    \caption{Question-answer relationships in 108 randomly selected samples from the FQuAD development set. In bold the elements needed for the corresponding reasoning, in italics the selected answer.}
    \label{tab:dataset_analysis_difficulty}
\end{table*}

%% file: content/dataset_evaluation.tex
We present the evaluation metrics for the FQuAD dataset.
First, altough the evaluation metrics remain essentially the same as for SQuAD, some modifications must be taken into account regarding the French nature of the dataset.
Second, we evaluate the human performance on the FQuAD development and test datasets.
Third, we compare the FQuAD1.1 and SQuAD1.1 development datasets with several metrics.

\subsection{Evaluation metrics}

The Exact Match (EM) and F1-score metrics are common metrics being computed to evaluate the performances of a model.
The former measures the percentage of predictions matching exactly one of the ground truth answers.
The later computes the average overlap between the predicted tokens and the ground truth answer.
The prediction and ground truth are processed as bags of tokens. 
For questions labeled with multiple answers, the F1 score is the maximum F1 over all the ground truth answers.

The evaluation process in \citep{rajpurkar-etal-2016-squad} for both the F1 and EM ignores some English punctuation, i.e. the \textit{a}, \textit{an}, \textit{the} articles.
In order to remain consistent with the former approach, the French evaluation process ignores the following articles: \textit{le, la, les, l', du, des, au, aux, un, une}.

\subsection{Human performance}

Similarly to SQuAD, human performances are evaluated on the development and test sets in order to assess how humans agree on answering questions.
This score gives a comparison baseline when assessing the performance of a model.
To measure the human performance, for each question, two of the three answers are considered as the ground truth, and the third as the prediction. 
In order not to bias this choice, the three answers are successively considered as the prediction, so that three human scores are calculated. 
The three runs are then averaged to obtain the final human performance.
Both the F1 and EM score are computed based on this setup.

The table \ref{tab:dataset_evaluation_human_performace} reports the results obtained for FQuAD1.0 and FQuAD1.1.
The human score on FQuAD1.0 reaches 92.1\% F1 and 78.4\% EM on the test set and
92.6\% and 79.5\% on the development set.
On FQuAD1.1, it reaches 91.2\% F1 and 75.9\% EM on the test set and
92.1\% and	78.3\% on the development set.
We observe that there is a noticeable gap between the human performance on FQuAD1.0 test dataset and the human performance on the new samples of FQuAD1.1 with 78.4\% EM score on the 2189 questions of FQuAD1.0 test set and 74.1\% EM score on the 3405 new questions of FQuAD1.1 test set.
As explained in section \ref{section:dataset_collection} we insisted in our annotation guidelines of FQuAD1.1 that the questions should be more difficult. 
This gap in human performance constitutes for us a proof that answering to FQuAD1.1 new questions is globally more difficult than answering to FQuAD1.0 questions, hence making the final FQuAD1.1 dataset even more challenging.

\begin{table}[ht]
    \centering
    \resizebox{0.47\textwidth}{!}{%
    \begin{tabular}{l c c}
        Dataset                         & F1 [\%]   & EM [\%]   \\
        \hline
            FQuAD1.0-test.              & 92.1      & 78.4      \\
            FQuAD1.1-test               & 91.2      & 75.9      \\
            "FQuAD1.1-test new samples" & 90.5      & 74.1      \\
        \hline
            FQuAD1.0-dev                & 92.6      & 79.5      \\
            FQuAD1.1-dev                & 92.1      & 78.3      \\
            "FQuAD1.1-dev new samples"  & 91.4      & 76.7      \\
        \hline
    \end{tabular}}
    \caption{Human Performance on FQuAD}
    \label{tab:dataset_evaluation_human_performace}
\end{table}

\subsection{Comparing FQuAD1.1 and SQuAD1.1}

The SQuAD1.1 dataset \citep{rajpurkar-etal-2016-squad} reports a human score for the test set equal to 91.2\% F1 and 82.3\% EM.
Comparing the English score with the French ones, we notice that they are the same in terms of F1 score but  differ by 6\% on the Exact Match.
This difference indicates a potential structural difference between FQuAD1.1 and SQuAD1.1. 
To better understand it we first compare the answer type distributions, then we compare the answer lengths for both datasets and finally we explore how the evaluation score varies with the answer length.

\paragraph{Answer type distribution}{
The comparison in answer type distribution between the FQuAD1.1 and SQuAD1.1 datasets are reported in table \ref{tab:answers_comparison_fquad_squad}.
For both datasets, the most represented answer type is \texttt{Common Noun} with FQuAD1.1 scoring 26.6\% and SQuAD1.1 scoring 31.8\%. 
The less represented ones are \texttt{Adjective} and \texttt{Other} which have a noticeable higher proportion for SQuAD1.1 than FQuAD1.1
Compared to SQuAD1.1, a significant difference exists on structured entities such as \texttt{Person}, \texttt{Location}, and \texttt{Other Numeric} where FQuAD1.1 consistently scores above SQuAD1.1 with the exception of the \texttt{Date} category where FQuAD scores less.
Based on these observations, it is difficult to understand the difference in human score between the two datasets.
}

\begin{table}[ht]
    \centering
    \resizebox{0.47\textwidth}{!}{%
    \begin{tabular}{l c c}
        Answer type             & FQuAD1.1 [\%] & SQuAD1.1 [\%]     \\
        \hline
            Common noun         & 26.6          & 31.8              \\
            Person              & 14.6          & 12.9              \\
            Other proper nouns  & 13.8          & 15.3              \\
            Location            & 14.1          & 4.4               \\
            Date                & 7.3           & 8.9               \\
            Other numeric       & 13.6          & 10.9              \\
            Verb                & 6.6           & 5.5               \\
            Adjective           & 2.6           & 3.9               \\
            Other               & 0.9           & 2.7               \\
        \hline
    \end{tabular}}
    \caption{Answer type comparison for the development sets of FQuAD1.1 and SQuAD1.1}
    \label{tab:answers_comparison_fquad_squad}
\end{table}

\paragraph{Answer length}{
To compare the answer lengths for the FQuAD1.1 and SQuAD1.1 datasets, we first remove every punctuation signs as well as respectively french words \textit{le, la, les, l', du, des, au, aux, un, une} and english words \textit{a, an, the}.
Then answers are split on white spaces to compute the number of tokens for each answer.
The results are reported in figure \ref{fig:answer_length_fquad_squad}.
It appears clearly that FQuAD answers are generally longer than SQuAD answers. 
Furthermore, to highlight this important difference it is interesting to realise that the average number of tokens per answer for SQuAD1.1 is equal to 2.72 while it is equal to 4.24 for FQuAD1.1.
This indicates that reaching a high Exact Match score on FQuAD is more difficult than on SQuAD.
}

\input{figures/answer_length_fquad_squad}

\paragraph{Human performance as a function of the answer length}{
To understand if the answer length can impact the difficulty of the Reading Comprehension task, we group question and answer pairs in FQuAD and SQuAD by the number of tokens for each answer.
The figure \ref{fig:human-perf-by-length} shows the human performance as a function of the answer length.
On one hand, it is straightforward to notice that the Exact Match quickly declines with an increasing answer length for both FQuAD and SQuAD.
On the other hand, the F1 score is a lot less affected by answer length for both datasets.
We conclude from these distributions that the difference in answers lengths between FQuAD and SQuAD may explain part of the difference in human performance regarding EM metric, while it does not seem to have an impact on human performance regarding F1 metric.
And indeed, human performance regarding F1 metric is very similar between FQuAD and SQuAD.
It is possible that these variations in answers lengths distributions are due to structural differences between French and English languages.
}

\input{figures/human-perf-by-length}

\paragraph{Number of answers per question}{
As indicated in \citep{rajpurkar-squad-v2}, the SQuAD1.1 and SQuAD2.0 development and test sets have on average 4.8 answers per question.
By means of comparison, the FQuAD1.1 datasets has on average 3 answers per question for the development and test sets.
The more answers to a question there are, the more likely it is that any other answer is equal to one of the expected answers. 
As a consequence, the higher number of answers in SQuAD1.1 contributes to the higher human performance compared to FQuAD1.1 regarding the exact match metric.
}

%% file: figures/answer_length_fquad_squad.tex
\pgfplotsset{compat = newest, legend style={at={(0.55,0.85)}, anchor=west}}

\begin{figure}[ht!]
    \centering
    \resizebox{0.45\textwidth}{!}{
    \begin{tikzpicture}[
        trim left=-0.3in, 
        trim right=\columnwidth-0.3in,
        scale=1.0
    ]
        \begin{axis}[
            xmin = 0, 
            xmax = 15,
            ymin = 0, 
            ymax = 40, 
            xtick distance = 2,
            ytick distance = 5,
            grid = both,
            minor tick num = 2, 
            major grid style = {lightgray},
            minor grid style = {lightgray!25},
            point meta=y,
            xlabel=Answer length,
            ylabel=Proportion \%,
        ]
            \addplot[curvered, mark=x, thick] coordinates {
              (1, 36.9)
              (2, 27.9)
              (3, 14.7)
              (4,  7.4)
              (5,  4.3)
              (6,  2.5)
              (7,  1.9)
              (8,  1.3)
              (9,  1)
              (10, 0.7)
              (11, 0.6)
              (12, 0.4)
              (13, 0.3)
              (14, 0.4)
            };
            \addplot[curveblue, mark= x, thick] coordinates {
              (1, 26)
              (2, 23.8)
              (3, 14.5)
              (4, 9.7)
              (5,  6.6)
              (6,  5)
              (7,  3.4)
              (8,  2.4)
              (9,  2.2)
              (10, 1.5)
              (11, 1.4)
              (12, 0.9)
              (13, 0.8)
              (14, 1.4)
            };
            \legend{SQuAD1.1, FQuAD1.1}
      \end{axis}
    \end{tikzpicture}}
    \caption{Answers lengths distribution for FQuAD and SQuAD}
    \label{fig:answer_length_fquad_squad}
\end{figure}
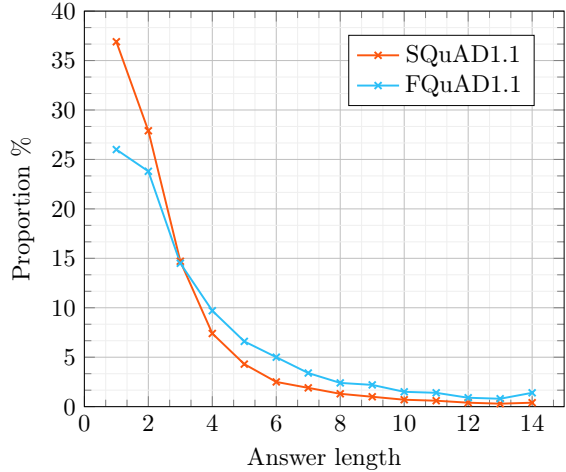

%% file: figures/human-perf-by-length.tex
\pgfplotsset{compat = newest, legend style={at={(0.10,0.25)},anchor=west}}

\begin{figure}[ht!]
    \centering
    \resizebox{0.45\textwidth}{!}{
    \begin{tikzpicture}[
        trim left=-0.3in, 
        trim right=\columnwidth-0.3in,
        scale=1.0
    ]
        \begin{axis}[
            xmin = 0, 
            xmax = 15,
            ymin = 0, 
            ymax = 100, 
            xtick distance = 2,
            ytick distance = 20,
            grid = both,
            minor tick num = 1, 
            major grid style = {lightgray},
            minor grid style = {lightgray!25},
            point meta=y,
            xlabel=length of answer in number of tokens,
            ylabel=human performance [\%],
        ]
            \addplot[color=red, mark=x, thick] coordinates {
              (1, 98.7)
              (2, 92.3)
              (3, 91)
              (4, 89.4)
              (5, 92.1)
              (6, 89.1)
              (7, 90)
              (8, 89)
              (9, 89.7)
              (10, 86.6)
              (11, 90.6)
              (12, 85.6)
              (13, 88.9)
              (14, 85.1)
            };
            \addplot[color=purple, mark=x, thick] coordinates {
              (1, 98.1)
              (2, 91.6)
              (3, 90.2)
              (4, 89.8)
              (5, 86.7)
              (6, 83.6)
              (7, 82.5)
              (8, 83.8)
              (9, 80.3)
              (10, 85.3)
              (11, 76.7)
              (12, 90.2)
              (13, 86.7)
              (14, 86.8)
            };
            \addplot[color=blue, mark= x, thick] coordinates {
              (1, 98.7)
              (2, 81.1)
              (3, 75.7)
              (4, 69.5)
              (5, 71.6)
              (6, 67)
              (7, 64.1)
              (8, 64.8)
              (9, 60.7)
              (10, 54.2)
              (11, 67)
              (12, 45.7)
              (13, 54.4)
              (14, 47.3)
            };
            \addplot[color=green, mark= x, thick] coordinates {
              (1, 98.1)
              (2, 82.1)
              (3, 77.4)
              (4, 71.7)
              (5, 65)
              (6, 58)
              (7, 61.3)
              (8, 50.6)
              (9, 55.4)
              (10, 49.3)
              (11, 50.9)
              (12, 35.7)
              (13, 50)
              (14, 60)
            };
            \legend{F1-FQuAD, F1-SQuAD, EM-FQuAD, EM-SQuAD}
      \end{axis}
    \end{tikzpicture}}
    \caption{Evolution of the F1 and EM human scores for the answers length of the development sets of FQuAD1.1 and SQuAD1.1}
    \label{fig:human-perf-by-length}
\end{figure}
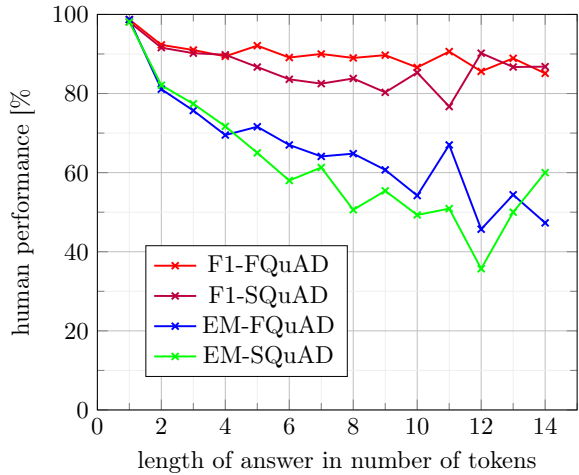

%% file: content/experiments.tex
We present the experiments that are carried out in order to evaluate both the quality of the new French Reading Comprehension dataset and the resulting fine-tuned models.
First we present the experimental set-up.
Second, the French monolingual language models and multilingual language models fine-tuning experiments are performed.
Finally, we investigate on one hand how zero-shot learning from English SQuAD1.1 performs on our dataset and on the other we evaluate the results with cross-lingual approaches based on the French translation of SQuAD1.1.

\subsection{Experimental set-up}

The experimental set-up is kept the same across all the experiments.
The number of epochs is set to $3$, with a learning rate equal to $3.0\cdot10^{-5}$.
The learning rate is scheduled according to a warm-up linear scheduler where the percentage ratio for the warm-up is consistently set to $6$\%.
The batch size is kept constant across the training and is equal to $8$ for the base models and $4$ for the large ones.
The optimizer that is being used is \texttt{AdamW} with its default parameters.
All the experiments were carried out with the HuggingFace transformers library \citep{Wolf2019HuggingFacesTS} on a single V100 GPU.

\subsection{Native French Reading Comprehension}

\paragraph{Monolingual vs. multilingual language models}{
The goal of these experiments is two fold.
First, we want to evaluate and compare how the French language models CamemBERT \citep{camembert} and FlauBERT \citep{flaubert} perform on FQuAD. 
Second, we want to evaluate how multilingual models perform when they are fine-tuned on the French dataset. 
For this purpose we train two multilingual models, i.e mBERT \citep{multilingual-bert} and the XLM-RoBERTa model \citep{xlmr}.
Finally, we will be able to compare the results for both the monolingual and multilingual models to understand how they perform on the French dataset.
Note that for each experiment, the fine-tuning is performed on the training sets of both FQuAD1.0 and FQuAD1.1 but are evaluated only on the development and test sets of FQuAD.1.1.
}

\paragraph{Performance analysis}{
An analysis of the predictions for the best trained model is carried out.
We have explored the distribution of answer and questions types in section \ref{section:dataset_analysis} and we report now the performance of the model in terms of F1 score and Exact Match for each category.
This analysis aims at understanding how the model performs on the various question and answer types.
}

\paragraph{Learning curve}{
The question of how much data is needed to train a question answering model remains relatively unexplored.
In our effort of annotating FQuAD1.0 and FQuAD1.1 we have consistently monitored the scores to know if the annotation process must be continued or stopped.
For this purpose, we present a learning curve obtained on the FQuAD1.1 test set by training CamemBERT\textsubscript{BASE} on an increasing number of question and answer samples.
Both the EM and F1 scores are reported on the learning curve.
}

\paragraph{PIAF}{
The French Dataset PIAF has been released after the first release of the present work.
In order to assess the impact of the PIAF released samples (3885 training samples), we perform two experiments using PIAF.
First, we evaluate the CamemBERT models fine-tuned on FQuAD1.0 on the new samples.
Second, we concatenate FQuAD1.0 and PIAF to train a new model and evaluate them on the test set of FQuAD1.1 to understand if the new samples bring additional score.
}

\subsection{Cross-lingual Reading Comprehension}

Cross-lingual Reading comprehension follows mainly two approaches as explained in \ref{section:related_work}.
On one hand, experiments carried out in \citep{mlqa} and \citep{xquad} evaluate how multilingual models fine-tuned on the English SQuAD1.1 dataset perform on other languages such as Spanish, Chinese or Arabic. 
On the other hand, initiatives such as \citep{spanishsquad} attempt to translate the dataset in the target language to fine-tune a model.
The newly obtained FQuAD dataset makes it now possible to test both approaches on the English-French cross-lingual set-up.
Note however that French is unfortunately not supported by the cross-lingual benchmark proposed by \citep*{mlqa}, \citep{xquad}.

First, we perform several experiments with a so called zero-shot learning approach.
In other words, we fine-tune several multilingual models on the English SQuAD1.1 dataset and we evaluate them on the FQuAD1.1 development set.
In addition to that, the opposite approach is also carried out, i.e. fine-tuned models on FQuAD1.1 are evaluated on the SQuAD1.1 development set.

Second, we fine-tune CamemBERT on the SQuAD1.1 training dataset translated into French.
For this purpose, the SQuAD1.1 training set is translated using NMT \citep{french-nmt}.
Note that the translation process makes it difficult to keep all the samples from the original dataset and, for the sake of simplicity, we discard the translated answers that do not align with the start/end positions of the translated paragraphs.
The resulting translated dataset \textit{SQuAD1.1-fr-train} contains about 40.7k question and answer pairs.
The fine-tuned model is then evaluated on the native French FQuAD1.1 development set.
This experiment helps us to understand how the translation process ultimately affects the performance of the model on native data rather than on the translated development set.




%% file: content/results.tex
In the present section, we present the results for the aforementioned evaluation experiments.
First, we present the results for the native French Reading Comprehension experiments along with the performance analysis for the best obtained model and a learning curve.
Second, we present the results for the cross-lingual Reading Comprehension experiments.

\subsection{Native French Reading Comprehension}

The training experiments on FQuAD1.1-train are summed up in table \ref{tab:experiments_large}, while training experiments on FQuAD1.0-train are summed up in table \ref{tab:experiments_base}.
The benchmark includes the training experiments for CamemBERT, FlauBERT, Multilingual BERT and XLM-R on training sets of FQuAD1.1 and FQuAD1.0
All the models are evaluated on the FQuAD1.1 test and development sets.

\begin{table}[ht]
    \small
    \centering
    \renewcommand{\arraystretch}{1.3}
    \resizebox{0.47\textwidth}{!}{%
    \begin{tabular}{l c c c c}
         & \multicolumn{2} {c} {FQuAD1.1-test} & \multicolumn{2} {c} {FQuAD1.1-dev}\\
         Model                  & F1    & EM    & F1    & EM\\
        \hline
        Human Perf.             & 91.2  & 75.9  & 92.1  & 78.3 \\
        \hline
        CamemBERT\textsubscript{BASE}
                                & 88.4  & 78.4  & 88.1  & 78.1 \\
        CamemBERT\textsubscript{LARGE}       
                                & \textbf{92.2} & \textbf{82.1} & \textbf{91.8} & \textbf{82.4} \\
        FlauBERT\textsubscript{BASE}
                                & 77.6  & 66.5  & 76.3  & 65.5 \\
        FlauBERT\textsubscript{LARGE}
                                & 80.5  & 69.0  & 79.7  & 69.3 \\
        mBERT  
                                & 86.0  & 75.4  & 86.2  & 75.5 \\
        XLM-R\textsubscript{BASE}
                                & 85.9  & 75.3  & 85.5  & 74.9 \\
        XLM-R\textsubscript{LARGE}
                                & 89.5  & 79.0  & 89.1  & 78.9 \\
        \hline
    \end{tabular}}
    \caption{Results of the experiments for various monolingual and multilingual models carried out on the training dataset of \textbf{FQuAD1.1-train} and evaluated on test and development sets of FQuAD1.1}
    \label{tab:experiments_large}
\end{table}

\begin{table}[ht]
    \small
    \centering
    \renewcommand{\arraystretch}{1.3}
    \resizebox{0.47\textwidth}{!}{%
    \begin{tabular}{l c c c c}
        & \multicolumn{2} {c} {FQuAD1.1-test} & \multicolumn{2} {c} {FQuAD1.1-dev} \\
         Model                  & F1    & EM    & F1    & EM \\
        \hline
        Human Perf.             & 91.2  & 75.9  & 92.1  & 78.3 \\
        \hline
        CamemBERT\textsubscript{BASE}        
                                & 86.0  & 75.8  & 85.5  & 74.1 \\
        CamemBERT\textsubscript{LARGE}
                                & \textbf{91.5} & \textbf{82.0} & \textbf{91.0} & \textbf{81.2} \\
        mBERT
                                & 83.9  & 72.3  & 83.1  & 71.8 \\
        XLM-R\textsubscript{BASE}
                                & 82.2  & 71.4  & 82.4  & 71.0 \\
        XLM-R\textsubscript{LARGE}
                                & 88.7  & 78.5  & 88.2  & 77.5 \\
        \hline
    \end{tabular}}
    \caption{Results of the experiments for various monolingual and multilingual models carried out on the training dataset of \textbf{FQuAD1.0-train} and evaluated on test and development sets of FQuAD1.1}
    \label{tab:experiments_base}
\end{table}

\paragraph{Monolingual models}
The CamemBERT\textsubscript{BASE} trained on FQuAD1.1 reaches 88.4\% F1 and 78.4\% EM as reported on \ref{tab:experiments_large}.
Interestingly, the base version surpasses the Human Score in terms of Exact Match on the test set.
The best model, CamemBERT\textsubscript{LARGE} trained on FQuAD1.1 reaches a performance of 92.2\% F1 and 82.1\% EM on the test set, which is the highest score across the experiments and surpasses already the Human Performance for both metrics on the test and development sets.
By means of comparison, the best model of the SQuAD1.1 leader-board reaches 95.1\% F1 and 89.9\% EM on the SQuAD1.1 test set \citep{xlnet}.
Note that while the size of FQuAD1.1 remains smaller than its english counterpart, the aforementioned results yield a very promising baseline.
Note further that the same model reaches a performance of 93.3\% F1 and 84.6\% EM on the test set of FQuAD1.0, hereby supporting the fact that FQuAD1.1 includes more difficult question \ref{section:dataset_evaluation}.
The FlauBERT\textsubscript{BASE} and FlauBERT\textsubscript{LARGE} model fine-tuned on the FQuAD1.1 training dataset yield a surprisingly low performance of respectively 77.6/66.5\% and 80.6/70.3\% F1/EM score. 
Indeed, it is reported that FlauBERT rivals or even surpasses CamemBERT performances on several downstream tasks such as Text Classification, Natural Language Inference (NLI) or Paraphrasing \citep{flaubert}.

\paragraph{Multilingual models}
The results of the experiments carried out for the multilingual models reported in \ref{tab:experiments_large} and \ref{tab:experiments_base} show that they perform also very well when evaluated on the test and development sets of FQuAD1.1.
The top performer in this category is XLM-R\textsubscript{LARGE} which reaches 89.5\% F1 and 79\% EM on FQuAD1.1-test.
The model XLM-R\textsubscript{BASE} scores 85.9\% F1 and 75.3\% EM on the test set.
Comparatively, mBERT model reaches a similar performance with 86.0\% F1 and 75.4\% EM.
These experiments show that monolingual language models reach stronger performances than multilingual models overall.
Nevertheless, it is important to note that XLM-R\textsubscript{LARGE} model performs better than CamemBERT\textsubscript{BASE} on both the test and development sets and even surpasses the Human Performance in terms of Exact Match on the test set.

\paragraph{Performance analysis}
Our best model CamemBERT\textsubscript{LARGE} is used to run the performance analysis on the question and answer types.
Tables \ref{tab:perfquestiontype} and \ref{tab:perfanswertype} present the results sorted by F1 score.
The model performs very well on structured data such as \texttt{Date}, \texttt{Numeric} or \texttt{Location}. 
Similarly, the model performs well on questions seeking for structured information, such as \texttt{How many}, \texttt{Where}, \texttt{When}.
The \texttt{Person} answer type human score is very high on EM metric, meaning that these answers are easier to detect exactly probably because the answer is in general short.
On the other end, the \texttt{How} and \texttt{Why} questions that probably expect a long and wordy answer are among the least well addressed. 
Note that \texttt{Verb} answers EM score is also quite low. 
This is probably due to either the variety of forms a verb can take, or to the fact that verbs are often part of long and wordy answers, which are by definition difficult to match exactly.
Some prediction examples are available in the appendix. 
Selected samples are not part of FQuAD, but were sourced from Wikipedia.

\begin{table}[ht]
    \small
    \centering
    \renewcommand{\arraystretch}{1.3}
    \resizebox{0.44\textwidth}{!}{%
    \begin{tabular}{l c c c c }
         Question Type & $F1$ & $EM$ & $F1_h$ & $EM_h$  \\
        \hline
        How many    & \textbf{96.3} & \textbf{87.8} & 93.3 & 82.1\\
        When        & 96.1 & 83.3 & 92.6 & 78.3\\
        Who         & 93.1 & 87.7 & 95.7 & 90.5\\
        Where       & 92.7 & 74.3 & 88.4 & 66.5\\
        What (que)  & 91.8 & 76.6 & 91.3 & 77.6\\
        Why         & 91.5 & 61.9 & 88.1 & 56.8\\
        What (quoi) & 89.8 & 64.9 & 88.3 & 66.1\\
        How         & 88.5 & 70.5 & 88.4 & 70.1\\
        Other       & 77.8 & 53.3 & 84.7 & 58.3\\
        \hline
    \end{tabular}}
    \caption{Performance on question types. $F1_h$ and $EM_h$ refer to human scores} 
    \label{tab:perfquestiontype}
\end{table}

\begin{table}[ht]
    \small
    \centering
    \renewcommand{\arraystretch}{1.3}
    \resizebox{0.47\textwidth}{!}{%
    \begin{tabular}{l c c c c }
        Answer Type & $F1$ & $EM$ & $F1_h$ & $EM_h$  \\
        \hline
        Date                & \textbf{95.8} & \textbf{82.1} & 92.6 & 78.1\\
        Other               & 94.6 & 75.6 & 84.4 & 63.7 \\
        Location            & 92.8 & 80.7 & 92.0 & 78.5 \\
        Other numeric       & 92.8 & 79.1 & 91.7 & 76.7 \\
        Person              & 92.5 & 80.8 & 93.4 & 82.6 \\
        Other proper nouns  & 92.5 & 78.3 & 91.9 & 78.0 \\
        Common noun         & 91.3 & 74.4 & 89.8 & 73.1 \\
        Adjective           & 89.6 & 73.1 & 90.8 & 71.6 \\
        Verb                & 88.5 & 58.7 & 87.7 & 60.9 \\
        \hline
    \end{tabular}}
    \caption{Performance on answer types. $F1_h$ and $EM_h$ refer to human scores} 
    \label{tab:perfanswertype}
\end{table}

\paragraph{Learning curve}
The learning curve is obtained by performing several experiments with an increasing number of question and answer samples randomly taken from the FQuAD1.1 dataset.
For each experiment, CamemBERT\textsubscript{BASE} is fine-tuned on the training subset and is evaluated on the FQuAD1.1 test set.
The F1 scores and Exact Match are reported on the figure \ref{fig:learningcurve} with respect to the number of samples involved in the training.
The figure shows that both the F1 and EM score follow the same trend.
First, the model is quickly improving upon the first 10k samples.
Then, F1 and EM are progressively flattening upon augmenting the number of training samples.
Finally, they reach a maximum value of respectively 88.4\% and 78.4\%.
The results show us that a relatively low number of samples are needed to reach acceptable results on the reading comprehension task. 
However, to outperform the Human Score, i.e. 91.2\% and 75.9 \%, a larger number of samples is required.
In the present case CamemBERT\textsubscript{BASE} outperforms the Human Exact Match after it us trained on 30k samples or more.

\input{figures/learningcurve.tex}

\paragraph{PIAF Dataset}


The experiments carried out on PIAF are reported in table \ref{tab:results_piaf}. 
To ease the comparison we also add the results from table \ref{tab:experiments_base}.
The results show that the F1 and EM performances reach a significantly lower level than on FQuAD1.1-test. One of the reasons for such a gap is the fact that the PIAF dataset does not include several answers per question as it is the case in SQuAD1.1 or in the present work.

\begin{table}[ht]
    \small
    \centering
    \renewcommand{\arraystretch}{1.3}
    \resizebox{0.45\textwidth}{!}{%
    \begin{tabular}{l c c c c}
        & \multicolumn{2} {c} {  PIAF  } & \multicolumn{2} {c} {FQuAD1.1-test}\\
         Training data                  & F1    & EM  & F1    & EM \\
        \hline
        FQuAD1.0 \textit{(1)}        
                                & 68.15 & 48.79 & 86.0  & 75.8 \\
        FQuAD1.0 \textit{(2)} 
                                & 74.43 & 54.39 & 91.5 & 82.0 \\
        FQuAD1.0 + PIAF  \textit{(1)}
                                & - & - & 86.8 & 76.2 \\
        \hline
    \end{tabular}}
    \caption{Results of the experiments for CamemBERT trained on \textbf{FQuAD1.0-train} and evaluated on PIAF.  \textit{(1)} has been trained with CamemBERT\textsubscript{BASE}, \textit{(2)} has been trained with CamemBERT\textsubscript{LARGE}.}
    \label{tab:results_piaf}
\end{table}

\subsection{Cross-lingual Reading Comprehension}

The results for the experiments on the cross-lingual set-up are reported in table \ref{tab:experiments_cross_lingual}. 
On one hand, the French monolingual models are fine-tuned on the French translated version of SQuAD1.1 and evaluated on the development set of FQuAD1.1.
On the other hand, multi-language models are fine-tuned respectively on SQuAD1.1 and FQuAD1.1 and then evaluated respectively on the development sets of FQuAD1.1 and SQuAD1.1 in order to evaluate the performance of zero-shot learning set-up.

\begin{table}[ht]
    \small
    \centering
    \renewcommand{\arraystretch}{1.3}
    \resizebox{0.47\textwidth}{!}{%
    \begin{tabular}{l l c c c c}
         &  & \multicolumn{2} {c} {SQuAD1.1-dev} & \multicolumn{2} {c} {FQuAD1.1-dev}\\
         Model              &  Train Dataset      & F1 [\%] & EM [\%] & F1 [\%] & EM [\%]   \\
        \hline
        Human Perf.         &               & 91     & 80.5 & 92.1  & 78.3 \\
        \hline
        \multirow{2}{*}{CamemBERT\textsubscript{BASE}} 
                            & FQuAD1.1      & -     & -     & 88.1  & 78.1 \\
                            & SQuAD1.1-fr   & -     & -     & 81.8  & 67.8 \\
                            & Augmented     & -     & -     & 88.3	& 78.0 \\
        \hline
        \multirow{2}{*}{CamemBERT\textsubscript{LARGE}} 
                            & FQuAD1.1      & -     & -     & 91.8  & 82.4 \\
                            & SQuAD1.1-fr   & -     & -     & 87.5  & 73.9 \\
                            & Augmented     & -     & -     & 91.2	& 81.6 \\
        \hline
        \multirow{2}{*}{XLM-R\textsubscript{BASE}} 
                            & FQuAD1.1      & 83.0  & 73.5  & 85.5  & 74.9  \\
                            & SQuAD1.1      & 88.1  & 80.9  & 81.4  & 68.4  \\
        \hline
        \multirow{2}{*}{XLM-R\textsubscript{LARGE}}
                            & FQuAD1.1      & 88.8  & 79.5  & 89.1  & 78.9  \\
                            & SQuAD1.1      & 90.7 & 83.4   & 86.1  & 73.2     \\
        \hline
    \end{tabular}}
    \caption{Results for the zero-shot learning experiments on the SQuAD1.1 and FQuAD1.1 development sets}
    \label{tab:experiments_cross_lingual}
\end{table}

\paragraph{Translated Reading Comprehension}
First, the results for CamemBERT\textsubscript{BASE} fine-tuned on the French translated version of SQuAD1.1. show a performance of 81.8\% F1 and 67.8\% EM as reported in \ref{tab:experiments_cross_lingual}.
Compared to CamemBERT\textsubscript{BASE} fine-tuned on FQuAD, this result is about 6.3 points less effective in terms of F1 score and even more important in terms of EM score, i.e. 10.3.
Second, the results for CamemBERT\textsubscript{LARGE} show an improved performance of 87.5\% F1 and 73.9\% EM.
Compared to the native version, this result is lower by 4.3 points in terms of F1 Score and 8.5 points in terms of EM. 
These experiments show therefore that models fine-tuned on translated data do not perform as well as when they are fine-tuned on native dataset.
This difference is probably explained by the fact that NMT produces translation inaccuracies that impact the EM score more than F1 score.
When we merge the native and the translated dataset into what we call the Augmented dataset, we do not observe a significant performance improvement. 
Interestingly, the CamemBERT\textsubscript{LARGE} model performs slightly worse when fine-tuned on translated samples.

\paragraph{Zero-shot learning}
To evaluate how multi-language models transfer on other languages similarly to \cite{mlqa} and \cite{xquad}, we report the results of our experiments with XLM-R\textsubscript{BASE} and XLM-R\textsubscript{LARGE} in \ref{tab:experiments_cross_lingual}.
We find that XLM-R\textsubscript{BASE} trained on FQuAD1.1 reaches 83.0\% F1 and 73.5 \% EM on the SQuAD1.1 dev set.
When trained on SQuAD1.1 it reaches 81.4\% F1 and 68.4\% EM on the FQuAD1.1 dev set.
Next, we find that XLM-R\textsubscript{LARGE} reaches 88.8\% F1 and 79.5\% on the SQuAD1.1 dev set when trained on FQuAD1.1 and 86.1\% F1 and 73.2\% EM on the FQuAD1.1 dev set when trained on SQuAD1.1.
The results show that the models perform very well compared to the results when trained on the native French and native English datasets.
Indeed, XLM-R\textsubscript{BASE} shows a drop of only 4.1\% and 6.5\% in terms of F1 and EM score on the FQuAD1.1 dev set when compared to the model trained on the native french samples.
And XLM-R\textsubscript{LARGE} show a drop on 3.0\% and 5.7\% in terms of F1 and EM score.
Note that the same relationship can be observed for the model trained on FQuAD1.1 and evaluated on SQuAD1.1 although the drop in performance is slightly less important.
Interestingly, the large models perform in general very well on the cross-lingual zero-shot set-up.

%% file: figures/learningcurve.tex
\pgfplotsset{compat = newest, legend style={at={(0.7,0.2)},anchor=west}}

\begin{figure}[ht!]
    \centering
    \begin{tikzpicture}[
        trim left=-0.3in, 
        trim right=\columnwidth-0.3in,
        scale=1.0
    ]
        \begin{axis}[
            xmin = 0, 
            xmax = 55,
            ymin = 0, 
            ymax = 100, 
            xtick distance = 10,
            ytick distance = 20,
            grid = both,
            minor tick num = 1, 
            major grid style = {lightgray},
            minor grid style = {lightgray!25}, 
            nodes near coords,
            point meta=y,
            xlabel=$\text{\#samples in training set} \; \cdot 10^3$,
            ylabel=score \% on test set,
        ]
            \addplot[curvered, mark=x, thick] coordinates {
              (2.5, 49)
              (5.0, 78)
              (10.0, 82)
              (15.0, 84)
              (20.0, 85.3)
              (30.0, 87.2)
              (40.0, 88.0)
              (50.0, 88.4)
            };
            \addplot[curveblue, mark= x, thick] coordinates {
              (2.5, 33)
              (5.0, 64)
              (10.0, 70)
              (15.0, 72)
              (20.0, 74.3)
              (30.0, 76.4)
              (40.0, 77.8)
              (50.0, 78.4)
            };
            \legend{F1, EM}
      \end{axis}
    \end{tikzpicture}
    \caption{Evolution of the F1 and EM scores for CamemBERT\textsubscript{BASE} depending on the number of samples in the training dataset}
    \label{fig:learningcurve}
\end{figure}
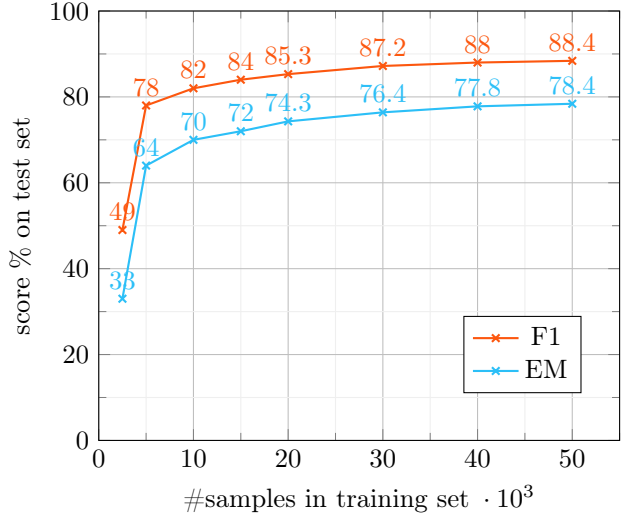

%% file: content/discussion.tex
The release of a native French Reading Comprehension dataset is motivated by the release of recent French monolingual models (\cite{camembert}, \cite{flaubert}) and by industrial opportunities.
In addition to that, we think that a French dataset opens up a wide range of possible experiments at the research level.
First, while it is generally accepted that monolingual models perform better than multilingual models we find that the gap is narrower than expected for the Reading Comprehension task.
Second, to fine-tune a model on a target language, translated datasets have been extensively used but the lack of native data to evaluate the approach, at least in French, makes it difficult to evaluate it.
Third, apart from Question Answering models for French applications, cross-lingual applications have found significant interest recently with \citep{xquad} and \citep{mlqa} where the need for quality annotated data on other languages than English are important to evaluate how models transfer across languages.

\subsection{Monolingual vs. multilingual language models}

Through our language models benchmark on FQuAD, we have evaluated several monolingual and multilingual models.
The CamemBERT\textsubscript{BASE} and CamemBERT\textsubscript{LARGE} models reach a very promising baseline and the large model even outperforms the Human Performance consistently across the development and test datasets.
Surprisingly we find very poor results for the FlauBERT\textsubscript{BASE} and FlauBERT\textsubscript{LARGE} models.

For comparable model sizes we find that the monolingual models outperform multilingual models on the Reading Comprehension task.
However, we find that multilingual models such as mBERT \citep{multilingual-bert} or XLM-R\textsubscript{BASE} and XLM-R\textsubscript{LARGE} \citep{xlmr} reach very promising scores.
We find that XLM-R\textsubscript{LARGE} performs consistently better than the monolingual model CamemBERT\textsubscript{BASE} on both the development and test sets of FQuAD1.1.
Let us further highlight that XLM-R\textsubscript{LARGE} reaches 79\% EM on FQuAD-test which is better than Human Performance, while the F1 score remains only 2\% below it.
As such a model is pre-trained on a multilingual corpus, we can hope that it could be used with reasonable performances on other languages. 

\subsection{Translated Reading Comprehension}

Fine-tuning CamemBERT\textsubscript{BASE} on a French translated dataset yields 81.8/67.8\% F1/EM on the FQuAD1.1 dev set.
By means of comparison, CamemBERT\textsubscript{BASE} scores 88.1/78.1\% F1/EM on the same set when trained with native French data.
We find here that there exists an important gap between both approaches.
Indeed, models that are fine-tuning on native data outperform models fine-tuned on translated data by an order of magnitude of 10\% for the Exact Match.

In \citep{spanishsquad}, the authors report a performance of 77.6/61.8\% F1/EM score when mBERT is trained on a Spanish-translated SQuAD1.1 and evaluated on XQuAD \citep{xquad}.
While the two approaches differ in terms of evaluation dataset, i.e. XQuAD is not a native Spanish dataset, and model, mBERT vs. CamemBERT, and although French and Spanish are different languages, they are close enough in their construction and structure, so that comparing these two approaches is relevant to us.
Given the level of effort put into the translation process in \citep{spanishsquad}, we think that both translation-based approaches, although using very recent language models, reach a performance ceiling with translated data.
We observe also that enriching native French training data with the translated samples does not improve the performances on the native evaluation set.
Given our experiments, we conclude therefore that there exist a significant gap between the native French and the French translated data in terms on quality and indicates that approaches based on translated data reach ceiling performances.

\subsection{Cross-lingual Reading Comprehension}

The zero-shot experiments show that multilingual models can reach strong performances on the Reading Comprehension task in French or English when the model has not encountered labels of the target language.
For example, the XLM-R\textsubscript{LARGE} model fine-tuned solely on FQuAD1.1 reaches a performance on SQuAD just a few points below the English Human Performance.
The same is also observed while fine-tuning solely on SQuAD1.1 and evaluating on the development set of FQuAD1.1.
We conclude here in agreement with \citep{xquad} and \citep{mlqa} that the transfer of models from French to English and vice versa relevant approach when no annotated samples are available in the target language.

The experiments also show that the zero-shot performances are better for SQuAD than for FQuAD.
This phenomenon can be explained by structural differences between French and English or an increased difficulty of FQuAD compared to SQuAD. 
It is also possible that the XLM-R language models used are capturing English language specifics better than for other languages because the dataset used for pre-training these models contains more English data.
Further experiments aiming at training multilingual models on both FQuAD1.1 and SQuAD1.1 may improve the results further.
This possibility is left for future works.

%% file: content/conclusion.tex
In the present work, we introduce the \textbf{F}rench \textbf{Qu}estion \textbf{A}nswering \textbf{D}ataset.
To our knowledge, it is the first dataset for native French Reading Comprehension. 
The contexts are collected from the set of high quality Wikipedia articles.
With the help of French college students, 60k+ questions have been manually annotated.
The FQuAD dataset is the result of two different annotation processes.
First, FQuAD1.0 is collected to build a 25k+ questions dataset.
Second, the dataset is enriched to reach 60k+ questions resulting in FQuAD1.1.
The development and test sets have both been enriched with additional answers for the evaluation process.

We find that the Human performances for FQuAD1.1 on the test and development sets reach respectively
a F1-score of 91.2\% and an Exact Match of 75.9\% and a F1-score of 92.1\% and an Exact Match of 78.3\%. 
Furthermore, we find that the Human performances on FQuAD1.1 reach comparable scores to SQuAD1.1.

Various experiments were carried out to evaluate the performances of fine-tuned monolingual and multilingual language models.
Our best model, CamemBERT\textsubscript{LARGE}, achieves a F1-score and an Exact Match of respectively 92.2\% and 82.1\%, surpassing the established Human performance in terms of F1-Score and Exact Match.
The experiments show that multilingual models reach promising results but monolingual models of comparable sizes perform better.

The FQuAD1.0 training and FQuAD1.1 development sets are made publicly available at \url{https://illuin-tech.github.io/FQuAD-explorer/} in order to foster research in the French NLP area.
The extension of the dataset to adversarial questions similarly to SQuAD2.0 is left for future works.



%% file: content/acknowledgement.tex
We would like to warmly thank Robert Vesoul, Co-Director of CentraleSupélec's Digital Innovation Chair and CEO of Illuin Technology, for his help and support in enabling and funding this project while leading it through.

We would also like to thank Enguerran Henniart, Lead Product Manager of Illuin annotation platform, for his assistance and technical support during the annotation campaign.

We more globally thank Illuin Technology for technical support, infrastructure and funding. 
We are also grateful to the Illuin Technology team for their reviewing and constructive feedbacks.

We share our warm thanks to Professor Céline Hudelot, professor at CentraleSupélec in charge of the Computer Science department, and head of the MICS laboratory (Mathematics and Informatics CentraleSupélec), for her guidance and support in this and other research works.  

Finally, we would also also like to thank Sebastian Ruder for his constructive feed-back on suggesting further experiments on the cross-lingual learning approach.

%% file: output_examples.tex
\section{Example model predictions}

\input{examples/fquad_prediction_example_brexit.tex}

\vspace{0.1in}
\input{examples/fquad_prediction_example_giec.tex}

%% file: examples/fquad_prediction_example_brexit.tex
\noindent\makebox[\textwidth][c]{%
\resizebox{0.85\textwidth}{!}{%
\fbox{\begin{minipage}{\textwidth}
    \textbf{Article:} Brexit
    
    \textbf{Paragraph:}
    \textit{\textcolor{red}{La possibilité d’un second référendum}} sur la question du \textit{\textcolor{orange}{projet de sortie du Royaume-Uni de l'Union européenne avait}} peu de chance de se réaliser avec le Premier ministre Boris Johnson. Elle fut toutefois fréquemment évoquée dans la presse britannique et étrangère. « Un second référendum est la seule façon de clore le débat » du Brexit a affirmé au journal Le Monde Tony Blair. Le député britannique Dominic Grieve expulsé du Parti conservateur avec \textit{\textcolor{blue}{21}} autres collègues en \textit{\textcolor{green}{septembre 2019}} pour avoir voté contre Boris Johnson afin de bloquer une sortie sans accord, a affirmé dans un entretien \textit{\textcolor{cyan}{à France 24 «}} que les Britanniques doivent connaître les conséquences d'un « no deal » » et va plus loin en affirmant : « je ne suis pas optimiste sur le fait qu’il soit possible de trouver un accord que le Parlement veuille. La seule solution est un second référendum. »

    \vspace{0.1in}
    
    \textbf{Question: }Quel évènement a été longuement mentionné dans la presse étrangère ?\\
    \textbf{Answer}: \textit{\textcolor{red}{La possibilité d’un second référendum}}
    
    \vspace{0.1in}
    
    \textbf{Question: }Combien de politiques ont été renvoyés du parti conservateur ?\\
    \textbf{Answer}: \textit{\textcolor{blue}{21}}
    
    \vspace{0.1in}
    
    \textbf{Question: }Sur quoi porte le second référendum ?\\
    \textbf{Answer}: \textit{\textcolor{orange}{projet de sortie du Royaume-Uni de l'Union européenne avait}}
    
    \vspace{0.1in}

    \textbf{Question: }Quel journal a accordé une interview à Dominic Grieve ?\\
    \textbf{Answer: } \textit{\textcolor{cyan}{à France 24 «}}
    
    \vspace{0.1in}
    
    \textbf{Question: }Quand Dominic Grieve a été renvoyé du parti conservateur ?\\
    \textbf{Answer: } \textit{\textcolor{green}{septembre 2019}}
    \vspace{0.1in}
\end{minipage}}}}

%% file: examples/fquad_prediction_example_giec.tex
\noindent\makebox[\textwidth][c]{%
\resizebox{0.85\textwidth}{!}{%
\fbox{\begin{minipage}{\textwidth}
    \textbf{Article:} Rapport du GIEC

    \textbf{Paragraph:}
    Le réchauffement planétaire atteindra les 1,5 \degree C entre \textit{\textcolor{red}{2030 et 2052}} si la température continue d'augmenter à ce rythme. Le RS15 (rapport spécial sur le réchauffement climatique de 1,5 \degree C) résume, d'une part, \textit{\textcolor{orange}{les recherches existantes sur l'impact qu'un réchauffement de 1,5 \degree C aurait sur la planète}} et, d'autre part, les mesures nécessaires pour limiter ce réchauffement planétaire.\\

    Même en supposant la mise en œuvre intégrale des mesures déterminées au niveau national soumises par les pays dans le cadre de l'Accord de Paris, les émissions nettes augmenteraient par rapport à 2010, entraînant un réchauffement d'environ 3 \degree C d'ici 2100, et davantage par la suite. En revanche, pour limiter le réchauffement au-dessous ou proche de 1,5 \degree C, il faudrait \textit{\textcolor{green}{diminuer les émissions nettes d'environ 45 \% d'ici 2030 et atteindre 0 \% en 2050.}} Même pour limiter le réchauffement climatique à moins de 2 \degree C, les émissions de CO2 devraient diminuer de 25 \% d'ici 2030 et de 100 \% d'ici 2075.\\

    Les scénarios qui permettraient une telle réduction d'ici 2050 ne permettraient de produire qu'environ 8 \% de l'électricité mondiale par le \textit{\textcolor{brown}{gaz}} et 0 à 2 \% par le charbon (à compenser par le captage et le stockage du dioxyde de carbone). Dans ces filières, les énergies renouvelables devraient fournir \textit{\textcolor{magenta}{70 à 85 \%}} de l'électricité en 2050 et la part de l'énergie nucléaire est modélisée pour augmenter. Il suppose également que d'autres mesures soient prises simultanément : par exemple, les émissions autres que le CO2 (comme le \textit{\textcolor{blue}{méthane, le noir de carbone, le protoxyde d'azote)}} doivent être réduites de manière similaire, la demande énergétique reste inchangée, voire réduite de 30 \% ou compensée par des méthodes sans précédentes d'élimination du dioxyde de carbone à mettre au point, tandis que \textit{\textcolor{cyan}{de nouvelles politiques et recherches}} permettent d'améliorer l'efficacité de l'agriculture et de l'industrie.

    \vspace{0.1in}
    
    \textbf{Question: }Quand risquons nous d'atteindre un réchauffement à 1.5 degrés? \\
    \textbf{Answer}: \textit{\textcolor{red}{entre 2030 et 2052}}
    
    \vspace{0.1in}
    
    \textbf{Question: }Quels sont les gaz à effet de serre autres que le CO2? \\
    \textbf{Answer}: \textit{\textcolor{blue}{méthane, le noir de carbone, le protoxyde d'azote)}}
    
    \vspace{0.1in}
    
    \textbf{Question: }Quelles recherches sont résumées dans ce rapport ?\\
    \textbf{Answer}: \textit{\textcolor{orange}{les recherches existantes sur l'impact qu'un réchauffement de 1,5 \degree C aurait sur la planète}}
    
    \vspace{0.1in}

    \textbf{Question: }Comment améliorer l'efficacité de l'industrie ? \\
    \textbf{Answer: } \textit{\textcolor{cyan}{de nouvelles politiques et recherches}}
    
    \vspace{0.1in}
    
    \textbf{Question: }Quelles sont les conséquences d'un scénario limitant le réchauffement à 1,5 degrés ?\\
    \textbf{Answer: } \textit{\textcolor{green}{diminuer les émissions nettes d'environ 45 \% d'ici 2030 et atteindre 0 \% en 2050.}}
    \vspace{0.1in}

    \textbf{Question: }Quelle part d'énergie doit être fournie par le renouvelable pour respecter l'accord ?\\
    \textbf{Answer: } \textit{\textcolor{magenta}{70 à 85 \%}}
    \vspace{0.1in}
    
    \textbf{Question: }Quelle source d'énergie sera limitée à une production de 8 \% si les émissions maximales sont respectées ? \\
    \textbf{Answer: } \textit{\textcolor{brown}{gaz}}
    \vspace{0.1in}

\end{minipage}}}}